%% file: ICRA2027_main.tex
\documentclass[letterpaper, 10pt, conference]{ieeeconf}

\IEEEoverridecommandlockouts
\usepackage{cite}
\usepackage{stfloats}
\usepackage{bm}

\usepackage{amsmath,amssymb}
\usepackage{graphicx}
\graphicspath{{figures/}}

\makeatletter
\let\NAT@parse\undefined
\makeatother
\usepackage[colorlinks=true,citecolor=green,linkcolor=black,urlcolor=green]{hyperref}
\newcommand{\R}{\mathbb{R}}
\newcommand{\Zset}{\mathcal{Z}}
\usepackage{microtype}

\usepackage{etoolbox}

\AtBeginEnvironment{equation}{\small}
\AtBeginEnvironment{equation*}{\small}
\AtBeginEnvironment{align}{\small}
\AtBeginEnvironment{align*}{\small}
\AtBeginEnvironment{gather}{\small}
\AtBeginEnvironment{gather*}{\small}
\AtBeginEnvironment{multline}{\small}
\AtBeginEnvironment{multline*}{\small}

\title{\LARGE\bf TRACER: Adaptive Multi-Robot Social Navigation via Joint Human-Response Prediction and Interaction-Aware Replanning}

\author{Lan Hu, Minghui Liwang, \textit{Senior Member, IEEE}, Wenbo Zhu, Xinlei Yi, \textit{Senior Member, IEEE}, Wei Gong,\\ \textit{Member, IEEE}, Yiguang Hong, \textit{Fellow, IEEE}, Seyyedali Hosseinalipour, \IEEEmembership{Senior Member, IEEE}
\thanks{L. Hu (2453781@tongji.edu.cn) is with the Guohao School, Tongji University, Shanghai, China. M. Liwang (minghuiliwang@tongji.edu.cn), W. Zhu (wbzhu@tongji.edu.cn), X. Yi (xinleiyi@tongji.edu.cn), W. Gong (weigong@tongji.edu.cn) and Y. Hong (yghong@tongji.edu.cn) are with the Shanghai Research Institute for Intelligent Autonomous Systems, the State Key Laboratory of Autonomous Intelligent Unmanned Systems, Department of Control Science and Engineering, Tongji University, Shanghai, China. S. Hosseinalipour (alipour@buffalo.edu) is with the Department of Electrical Engineering, University at Buffalo-SUNY, USA. Corresponding author: M. Liwang.}
\vspace{-5mm}}

\begin{document}

\maketitle
\thispagestyle{empty}
\pagestyle{empty}

\begin{abstract}
Multi-robot navigation in human-shared spaces is inherently interactive: coordinated robot motions influence how nearby entities respond, while those responses provide valuable information for subsequent robot decisions. However, existing methods typically address action-conditioned prediction, multi-robot planning, or online adaptation separately, and therefore lack a unified mechanism for modeling joint robot-entity interactions and adapting future decisions from executed interaction outcomes. To address this gap, we propose $\mathsf{TRACER}$, a bi-directional receding-horizon framework that closes the loop between prediction and adaptation. $\mathsf{TRACER}$ evaluates candidate (i.e., alternative feasible future motion plans for the robot team) trajectories using a per-entity probabilistic response model that separates individual-robot effects from non-additive pairwise interactions; after executing the selected trajectory prefix, it updates persistent identity-bound beliefs over latent response modes using the synchronized observed responses. These updated beliefs then guide subsequent candidate evaluation under probabilistic safety and response-aware cost criteria. 
Experiments show that \textit{(i)} $\mathsf{TRACER}$ more accurately captures non-additive multi-robot interaction effects than a capacity-matched additive predictor, \textit{(ii)} persistent identity-consistent evidence improves response prediction and downstream replanning, and \textit{(iii)} the complete $\mathsf{TRACER}$ framework improves collision-free completion over an independent-robot baseline on the SocialGym2 multi-robot social-navigation benchmark.
\end{abstract}


\vspace{-1.5mm}
\section{Introduction}
\vspace{-0.5mm}
As intelligent embodied agents move from closed, structured environments into large-scale human-shared spaces, a foundational assumption of navigation begins to break down \cite{r1}: humans are no longer merely exogenous elements of the environment governed by a fixed behavioral distribution; instead, \textit{they continuously adapt to how embodied agents move, interact, and are deployed at scale}~\cite{r3,r4,r5}. Through repeated long-term interaction, such adaptation may reshape walking habits, yielding conventions, and even perceived notions of safety and risk~\cite{r6}. Agents will, in turn, adapt to these emerging human responses, creating a closed-loop process of bi-directional co-evolution. Future embodied navigation  should therefore \textit{model how human and agent behaviors mutually reshape one another over time, and design navigation policies that actively steer this evolving interaction toward coordinated, stable, and socially compatible equilibria}.

To study these interactions, we consider human-shared environments with two agent classes:
\textit{robots}, whose motions are planned by the navigation system, and \textit{dynamic entities}, such as pedestrians and human-driven vehicles, whose motions are not controllable but may respond to robots' behaviors~\cite{peroi}. In this environment, multiple robots coexist in the same space, and thus an entity may react to their joint motion rather than to any single robot. We subsequently formulate multi-robot navigation as a closed-loop process that evaluates candidate trajectories through predicted entity responses, executes a short prefix of the selected plan, and updates the entity-specific response beliefs based on the observed responses.

\subsection{Motivation}\label{sec:motivation}
\vspace{-0.5mm}
Multi-robot navigation in human-shared environments hinges on capturing how coordinated robot decisions shape entity responses, and leveraging this coupling to navigate efficiently. Existing literature in this domain address action-conditioned and future-aware prediction~\cite{r3,r4,r5}, multi-robot planning and generative models~\cite{r7,r8}, and online belief updates for interacting agents~\cite{r9,r10}, but largely in isolation. Our setting instead requires these capabilities to operate jointly in a closed-loop: robot plans influence entity responses, while observed responses inform subsequent decisions. This setting motivates addressing the following research questions (RQs):

\noindent
$\bullet$ \textit{RQ1: How does the joint motion of multiple robots influence an entity's response beyond the effects of individual robots?} Existing action-conditioned predictors~\cite{r3,r4,r5,r22} typically condition on a single robot or a fixed participant set~\cite{r11,r13}, limiting their ability to compare how an entity may respond to alternative multi-robot plans. To address this, we represent each alternative as a \textit{candidate joint trajectory (CJT)}, comprising the planned trajectories of all active robots, and condition a per-entity probabilistic response model on this CJT. This formulation supports variable robot counts and orderings while enabling us to test whether joint robot motions induce non-additive response effects beyond those attributable to individual robots.

\noindent
$\bullet$ \textit{RQ2: How can observed robot-entity interaction outcomes be used to adapt subsequent multi-robot decisions?} RQ1 produces a response distribution for each CJT, whereas only the response observed under the executed plan provides evidence about the entity's underlying response behavior. Since decisions are repeatedly replanned over successive horizons, this evidence should be retained across planning windows rather than discarded or confused with counterfactual predictions. This raises two complementary requirements for belief maintenance: the system should preserve evidence accumulated from past executed interactions, while excluding predictions associated with plans that were never executed. Resetting the belief would violate the first requirement by discarding interaction history~\cite{r9,r10}, whereas incorporating predictions from unexecuted CJTs would violate the second by mixing counterfactual outcomes with observed evidence. Accordingly, we maintain an identity-bound posterior over response modes that is updated only from the executed robot prefix and the observed entity response. The updated posterior then evaluates CJTs in subsequent decision windows.

\begin{figure}[t]
  \centering
  \includegraphics[width=\columnwidth]{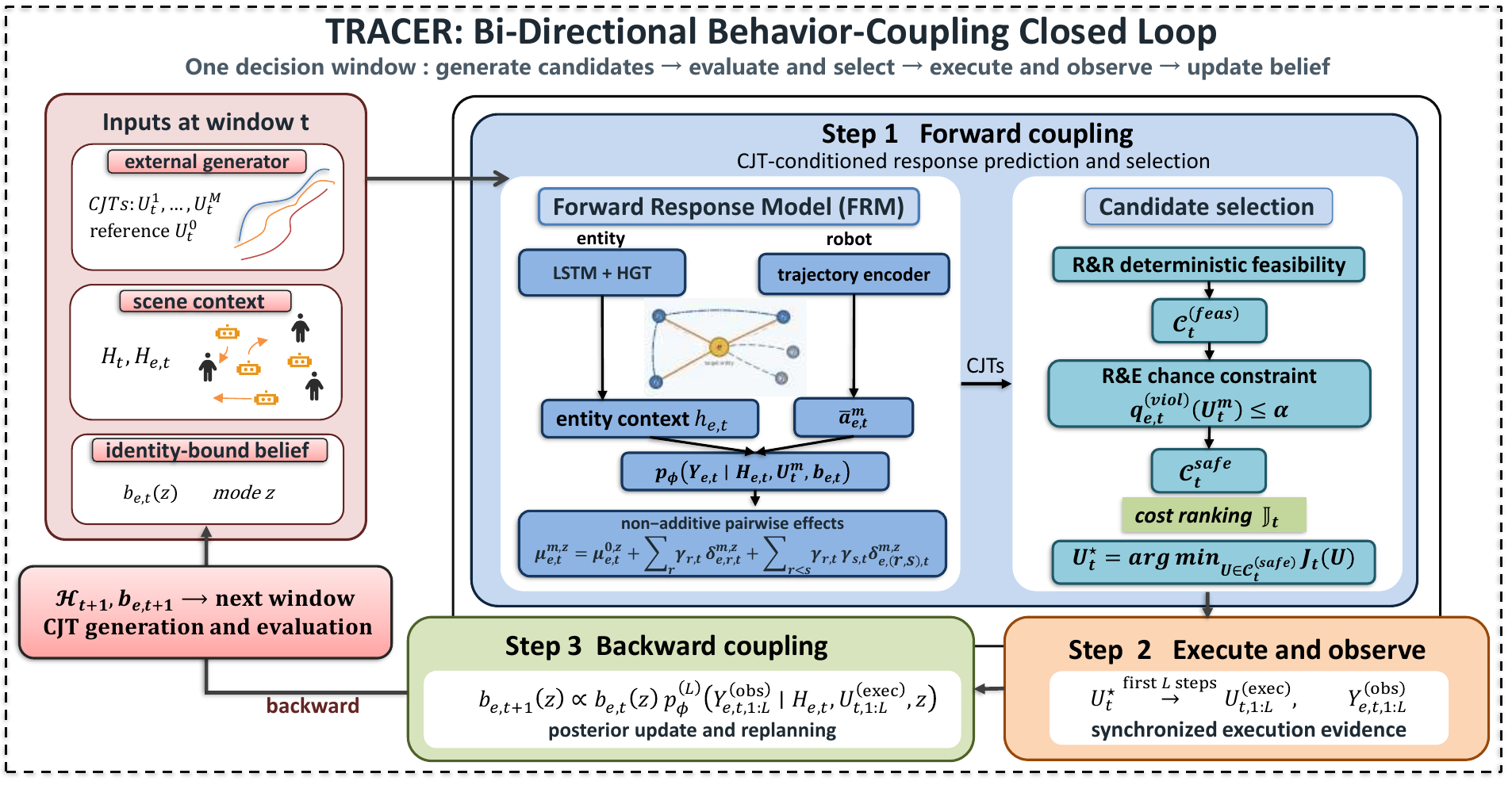}
  \vspace{-7.75mm}
 \caption{Overview of $\mathsf{TRACER}$: CJTs are generated externally; the forward response model predicts per-entity responses for every candidate; safety filtering and cost ranking select the CJT to execute; only the executed prefix and the observed real responses update the identity-bound mode posterior that shapes the next window's decisions.}
  \label{fig:framework}
  \vspace{-7mm}
\end{figure}

Motivated by the above RQs, we present $\mathsf{TRACER}$, a \textit{\underline{t}rajectory-conditioned \underline{r}esponse \underline{a}daptation for \underline{c}losed-loop \underline{e}ntity-aware \underline{r}eplanning for multi-robot social navigation} (illustrated in Fig.~\ref{fig:framework}). Our key contributions are as follows.

\noindent$\bullet$ \textit{Bi-Directional Behavior-Coupled Formulation:} We formulate multi-robot navigation as a closed-loop process in which CJTs induce per-entity response distributions, while executed interactions update identity-bound posteriors that shape future candidate selection. Our formulation further incorporates both deterministic and probabilistic safety constraints.

\noindent$\bullet$ \textit{CJT-Conditioned Forward Response Model:} Given a candidate joint trajectory, our proposed model predicts how each entity may respond while supporting variable numbers and orderings of robots. More importantly, it separates the effect of each robot from the combined response caused by robots acting jointly, allowing us to identify when multi-robot interactions produce non-additive behavioral effects.

\noindent$\bullet$ \textit{Evidence-Driven Backward Update:} After executing a selected CJT, we use the realized robot motion together with the entity’s observed response to refine an identity-specific belief. This updated belief is then carried into the next decision window, allowing future response prediction, safety assessment, and CJT selection to adapt to what was actually observed rather than to counterfactual predictions.

\noindent$\bullet$ \textit{Experimental Validation:} 
Controlled studies demonstrate the benefits of our joint-response modeling and identity-consistent evidence for prediction and replanning. Evaluations further support the practical utility of our method, with $\mathsf{TRACER}$ improving collision-free completion over the independent-robot $\mathsf{GoAlone}$ baseline on SocialGym2.

\subsection{Literature Review}\label{sec:related}
This section reviews the related works, summarizes key findings, and highlights their main differences from $\mathsf{TRACER}$.

~\noindent\textit{(i) Action-Conditioned Response Prediction:}
Existing approaches model how dynamic entities respond to robot motion using either hand-crafted interaction rules or learned policies. For example, rule-based social-force and reciprocal-avoidance models encode interaction through hand-designed motion rules, while learned collision-avoidance policies map local observations to actions~\cite{r16,r17}. Recent probabilistic forecasting methods model multimodal scene evolution and dependencies among multiple agents~\cite{r11,r13,r20,jppd}. A complementary direction explicitly conditions predicted responses on a planned robot trajectory through conditional simulation, generative prediction, bilevel model predictive control (MPC), and future-aware world models~\cite{r3,r4,r5,r22,sicnav_diffusion}. These methods establish the forward connection from robot actions to entity responses, but they typically condition on a single robot trajectory, a fixed participant set, or a scene-level distribution. As a result, they do not directly support per-entity comparison of alternative variable-size multi-robot CJTs. Moreover, while general interaction-analysis and high-dimensional effect-decomposition methods provide tools for separating joint effects~\cite{r27}, they have not been developed for our robot-entity navigation setting of interest.

~\noindent\textit{(ii) Joint Multi-Robot Decision Making:}
Multi-robot navigation methods coordinate robots while accounting for crowd interactions, collision avoidance, and deadlock prevention through game-theoretic planning, learned human models, control barrier functions, partial-belief planning, or scene generation~\cite{r7,r8,r25,r31,r32}. These approaches typically produce joint controls, navigation policies, or scene-level predictions as part of the planning process. In contrast, $\mathsf{TRACER}$ does not prescribe how CJTs are generated. Instead, it treats the candidate generator as an external module and evaluates each supplied CJT through per-entity response prediction, after which the resulting predictions are used for safety filtering and candidate ranking. This separation decouples response modeling from multi-robot trajectory generation, making our framework independent of any specific candidate-generation mechanism (i.e., any multi-robot planner/trajectory sampler).

~\noindent\textit{(iii) Adaptation from Executed Interaction:}
Interactive planning methods often infer latent quantities such as goals, rewards, driving styles, or game parameters from observed interactions~\cite{r6,r9,r10,r33,r35,r36,r37,r38,r41}. Related active-probing and dual-control approaches additionally balance task progress with information acquisition~\cite{r39,r40}. These works establish the importance of using interaction outcomes to refine future decisions, but they do not generally couple three elements required in our setting: a persistent identity-bound posterior, updates based only on the executed robot motion and synchronized entity response, and posterior-guided evaluation of alternative CJTs in subsequent decision windows.

\section{Problem Formulation}\label{sec:formulation}
We study multi-robot navigation under bi-directional human--robot interaction: each robot moves toward its destination subject to dynamics, robot--robot ($\mathsf{R\&R}$) coordination, and probabilistic robot--entity ($\mathsf{R\&E}$) safety constraints. At each decision window, the system evaluates multiple CJTs using predicted entity responses, executes a short prefix of the selected CJT, and updates the response posterior from the observed interaction, where the \textit{forward coupling} captures how a CJT shapes entity responses, while the \textit{backward coupling} captures how execution evidence influences subsequent CJT selection. We next describe our system's components.
\vspace{-2mm}

\subsection{Scene and Candidate Trajectories}\label{sec:scene}
In a two-dimensional (2D) shared space, we consider two agent types at each decision window $t$: robots $\mathcal{R}_t$ and dynamic entities $\mathcal{E}_t$. Their counts, $|\mathcal{R}_t|$ and $|\mathcal{E}_t|$, may vary as agents enter or leave the considered region. Each robot $r \in \mathcal{R}_t$ can be directly controlled and navigates toward a destination $g_r$; whereas each entity $e \in \mathcal{E}_t$ is an independently acting participant with its own behavioral intent, whose motion may adapt in response to the robots and influence subsequent planning. We next present a set of key definitions:

\noindent\textbf{Definition 1 (Joint Trajectories).}
For an active robot set $\mathcal R_t$, a \textit{joint trajectory (JT)} over the $T$-step horizon refers to a synchronized tuple $U_t=(\tau_{r,t})_{r\in\mathcal R_t}$, where $\tau_{r,t}$ captures the time-ordered sequence of position-velocity states for robot $r$. At the beginning of window $t$, an upstream candidate generator, such as a multi-robot planner or trajectory sampler, returns $M$ such joint trajectories for evaluation, indexed by $m=1,\ldots,M$. Each JT with index $m$, denoted by $U_t^m=(\tau_{r,t}^m)_{r\in\mathcal R_t}$ is called a \emph{candidate joint trajectory (CJT)}. Each CJT defines an \textit{alternative} prior to execution, with only $U_t^\star$ executed. The fixed reference JT $U_t^0$ serves as the baseline for response deviations, independently of the candidate set.

\noindent
\textbf{Definition 2 (Evaluated Entity Subset).}
We denote the evaluated subset as $\mathcal{E}^{(\mathrm{eval})}_t \subseteq \mathcal{E}_t$, which encompasses all entities that may enter the $\mathsf{R}\&\mathsf{E}$ safety distance under at least one CJT within the $T$-step prediction horizon. Only entities in $\mathcal{E}^{(\mathrm{eval})}_t$ are subject to candidate-level safety constraints.

\noindent
\textbf{Definition 3 (Global and Entity-local Context).}
The global context $\mathcal{H}_t$ comprises the information available to the candidate generator at the beginning of $t$, including the recent states of robots and entities, the locally feasible space, and robot destinations. Also, for each entity $e$, we denote its local context as $\mathcal{H}_{e,t}$, which contains the information used by the response model, including a fixed-length recent motion history up
to the beginning of window $t$, nearby entities, and the robots' relative geometry and motion.

At each decision window $t$, the system evaluates a set of CJTs and executes the first $L$ steps of the selected trajectory, where $1 \leq L < T$, before replanning from newly synchronized observations. We refer to these $L$ executed steps as the \textit{executed prefix} and the remaining $T-L$ steps as the \textit{unexecuted suffix}. To process a variable number of active robots with one model, each CJT
defined above is placed in $R^{(\max)}$ input slots (i.e., fixed input positions, each reserved for one robot). Let
$\gamma_{r,t}\in\{0,1\}$ indicate whether slot $r$ contains an active
robot, we have:
\vspace{-2mm}

\begin{equation}
|\mathcal R_t|
=
\sum_{r=1}^{R^{(\max)}}\gamma_{r,t},
\quad
r=1,\ldots,R^{(\max)} .
\label{eq:slot_mask}
\vspace{-2mm}
\end{equation}

For CJT $U^m_t $, slot $r$ carries $\tau_{r,t}^{m}$ when
$\gamma_{r,t}=1$; otherwise it is padded and masked. Since the slot index serves only as an input placeholder, permuting the active robots, their states, and their associated candidate trajectories does not change the CJT representation or the predicted entity response.

Each CJT is checked for deterministic
robot-side feasibility. Let $\mathsf{Dyn}(\cdot)$, $\mathsf{Obs}(\cdot)$, and
$\mathsf{RR}(\cdot)$ denote Boolean checks for robot dynamics,
static-obstacle avoidance, and $\mathsf{R\&R}$ spatio-temporal
non-overlap. We define the robot-feasible set as
\vspace{-7.5mm}

\begin{equation}
\mathcal{C}_t^{(\mathrm{feas})}
=
\left\{
U_t^m\,\middle|\,
\mathsf{Dyn}(U_t^m)\land
\mathsf{Obs}(U_t^m)\land
\mathsf{RR}(U_t^m)
\right\}.
\label{eq:feas}
\vspace{-1.5mm}
\end{equation} 
In contrast, $\mathsf{R\&E}$ safety cannot be verified deterministically due to uncertainty in future entity motion and is therefore imposed subsequently as a per-entity probabilistic constraint.

\subsection{Response Modes and Evidence}\label{sec:modes}
For entity $e$ in decision window $t$, let
$Y_{e,t}\in\mathbb{R}^{T\times 2}$ denote its velocity response over the
next $T$ prediction steps, with row $\ell$ giving the 2D velocity at
step $\ell$ $(\ell=1,\ldots,T)$. Each entity has a latent \textit{response mode}
$z_e\in\Zset,\Zset=\left\{1,\ldots,|\Zset|\right\}$ (at test time, the mode is latent and inferred from executed interactions). 
For a past window $t'<t$, the execution evidence associated with entity $e$ is given by
\vspace{-3mm}

\begin{equation}
\hspace{-1mm}
\resizebox{0.45\textwidth}{!}{$
\xi_{e,t'} {=} \left(\mathcal H_{e,t'},\ U_{t',1:L}^{(\mathrm{exec})},\ Y_{e,t',1:L}^{(\mathrm{obs})}\right),~
\mathcal D_{e,<t} {=} \left\{\xi_{e,t'}:t'<t\right\},
$}
\label{eq:evidence}
\hspace{-2mm}
\vspace{-2mm}
\end{equation}
where $U_{t',1:L}^{(\mathrm{exec})}$ is the measured robot position--velocity prefix that was actually executed, and $Y_{e,t',1:L}^{(\mathrm{obs})}$ is the synchronized entity response over the same $L$ steps. Predictions for unexecuted CJTs, and that for the unexecuted suffix of the selected CJT, are counterfactual and do not enter $\mathcal D_{e,<t}$. The identity-bound posterior at the beginning of $t$ is given by
\vspace{-2mm}

\begin{equation}
b_{e,t}(z) = \Pr\left(z_e=z\mid\mathcal D_{e,<t}\right),
\label{eq:posterior}
\vspace{-2mm}
\end{equation}
where $z\in\Zset$ and $\sum_{z\in\Zset}b_{e,t}(z)=1$.
Assuming that the current response is conditionally independent of earlier evidence given the current entity-local context, the predictive distribution for CJT $U^m_t$ is given by
\vspace{-4mm}

\begin{equation}
\hspace{-2mm}
\resizebox{0.46\textwidth}{!}{$p_{\phi}\left(Y_{e,t}\mid \mathcal H_{e,t},U^m_t,b_{e,t}\right){=}\sum_{z\in\Zset}b_{e,t}(z)\,p_{\phi}\left(Y_{e,t}\mid \mathcal H_{e,t},U^m_t,z\right)\hspace{-.7mm}.
$}
\label{eq:mixture}
\hspace{-2mm}
\end{equation}
Thus, changing $U^m_t$ changes the mode-conditional response distributions in \eqref{eq:mixture}, while executed evidence changes their mixture weights through $b_{e,t}$.

\subsection{Navigation Objective}\label{sec:decision}
We next formulate  multi-robot
navigation in human-shared spaces as a bi-directional
behavior-coupled, belief-conditioned, receding-horizon problem given by:
\vspace{-1.8mm}

\begin{equation}
\begin{aligned}
\bm{\mathcal{P}}:
&\arg\min_{U\in \mathcal{C}_t^{(\mathrm{feas})}}\mathbb{J}_t\\
\text{s.t.}\quad
&q_{e,t}^{(\mathrm{viol})}
\left(U\mid \mathcal H_{e,t},b_{e,t}\right)
\leq\alpha,\quad
\forall e\in\mathcal E_t^{(\mathrm{eval})},
\end{aligned}
\label{eq:objective}
\vspace{-2mm}
\end{equation}
where $\mathbb{J}_t=\lambda^{(\mathrm{goal})}J^{(\mathrm{goal})}_t
+\lambda^{(\mathrm{risk})}J^{(\mathrm{risk})}_t
+\lambda^{(\mathrm{brdn})}J^{(\mathrm{brdn})}_t$ balances goal
progress, residual interaction risk, and response burden objectives, the exact expressions of which are later derived in Section~\ref{sec:method}, where 
coefficients
$\lambda^{(\mathrm{goal})},\lambda^{(\mathrm{risk})},
\lambda^{(\mathrm{brdn})}\geq0$ weight these objective terms. Notably, $U_t^\star$ is obtained by
solving $\bm{\mathcal{P}}$; after executing its first $L$ steps, the
measured robot prefix and synchronized entity responses are added to
\eqref{eq:evidence}, and the updated posteriors are used in the next
decision window. In this problem, the forward coupling is realized by predicting each
entity's response under every CJT, whereas the backward coupling is
realized by updating its response belief from the executed interaction. In essence, at each decision window $t$, each CJT is represented with the active-slot mask in \eqref{eq:slot_mask} and restricted to the robot-feasible set in \eqref{eq:feas}. The selected CJT advances each robot toward its own destination while maintaining
$\mathsf{R\&E}$ safety and limiting unnecessary entity motion; after execution,
the measured prefix and synchronized entity response are recorded as the
evidence in \eqref{eq:evidence}, yielding the posterior in \eqref{eq:posterior}
for subsequent CJT evaluation.
Section~\ref{sec:method} details our method, called $\mathsf{TRACER}$, to solve $\bm{\mathcal{P}}$.

\section{Methodology}\label{sec:method}

\subsection{Overview of the Bi-Directional Coupling Mechanism}\label{sec:overview}

$\mathsf{TRACER}$ operates through two coupled mechanisms at each decision window: a forward coupling from candidate robot motions to predicted entity responses and a backward coupling from executed interactions to updated response beliefs. In the forward coupling, the forward response model (FRM) evaluates the mode-conditional response distribution
$p_{\phi}(Y_{e,t}\mid \mathcal H_{e,t},U_t^m,z)$
for each supplied CJT and response mode. These predictions, weighted by the current identity-bound posterior $b_{e,t}$ in \eqref{eq:mixture}, are then used for probabilistic safety filtering and cost-based selection of the CJT to execute (Section~\ref{sec:forward}). In the backward coupling, the executed robot prefix and synchronized entity response provide evidence for updating $b_{e,t}$ through an evidence-driven Bayesian update (EBU) (Section~\ref{sec:backward}). The resulting posterior is carried into the next decision window, where it influences the evaluation and selection of new CJTs. Thus, the forward coupling addresses RQ1 in Section I-A by modeling how alternative joint robot motions affect entity responses, while the backward coupling closes the loop required by RQ2 by adapting subsequent decisions to observed interactions. Overall, each decision window follows three steps: evaluate and select a CJT, execute a prefix and record the resulting interaction, and update the identity-bound posterior. 

\subsection{Forward Coupling: From CJTs to Selected Plan}\label{sec:forward}
The forward coupling turns a set of CJTs into a
selected plan by predicting each entity's response under every CJT and
using those predictions for safety filtering and cost ranking. It consists of a FRM and a selection rule, detailed below.

\subsubsection{Forward Response Model (FRM)}\label{sec:frm}
For each evaluated entity $e$, the FRM builds a local heterogeneous
interaction graph $G_{e,t}=(V_{e,t},E_{e,t})$ from $\mathcal H_{e,t}$, with $e$ as the
target node and with nearby entities and active robots as additional nodes in $V_{e,t}$.
The edges in $E_{e,t}$ have types $\mathsf{R\&R}$, $\mathsf{R\&E}$, and
entity--entity ($\mathsf{E\&E}$). For node $i\in V_{e,t}$, let
$s_{i,k}=p_{i,k}\in\mathbb{R}^2$ be its 2D position and
$s_{i,t-K_h+1:t}=(s_{i,k})_{k=t-K_h+1}^{t}$ its $K_h$-step history. A shared LSTM encodes each node's
history, and two heterogeneous graph transformer (HGT) layers~\cite{r43} propagate messages by edge type. In parallel, a trajectory encoder represents how each robot $r$ may move under candidate $m$ using its relative state $x_{e,r,t}$ and candidate trajectory $\tau^m_{r,t}$. We denote the shared LSTM encoder, the two successive HGT layers, and the trajectory encoder by $\psi^{(\mathrm{LSTM})}$, $\psi^{(\mathrm{HGT})}_1$ and $\psi^{(\mathrm{HGT})}_2$, and $\psi^{(\mathrm{traj})}$, respectively:
\vspace{-1.8mm}

\begin{equation}
\begin{aligned}
h^{(0)}_{i,t} &= \psi^{(\mathrm{LSTM})}(s_{i,t-K_h+1:t}),\quad i\in V_{e,t},\\
h_{e,t} &= \psi^{(\mathrm{HGT})}_2\!\circ\!\psi^{(\mathrm{HGT})}_1\!\left(G_{e,t},\{h^{(0)}_{i,t}\}_{i\in V_{e,t}}\right)_e,\\
a^m_{e,r,t} &= \psi^{(\mathrm{traj})}\left(x_{e,r,t},\tau^m_{r,t}\right).
\end{aligned}
\label{eq:hist}
\vspace{-1.7mm}
\end{equation}

\noindent Here, $h_{e,t}$ is the scene representation, and $a^m_{e,r,t}$ is the encoded candidate
trajectory. These are aggregated and combined with the mode and cardinality
embeddings $u(z)$, $u(|\mathcal R_t|)$ to obtain
\vspace{-9mm}

\begin{equation}
\hspace{-2mm}
\bar{a}^m_{e,t} = \tfrac{1}{|\mathcal R_t|}\sum_{r=1}^{R^{(\max)}}\gamma_{r,t}\,a^m_{e,r,t},~~~
q^z_{e,t} = \left[h_{e,t};\ u(z);\ u(|\mathcal R_t|)\right],
\label{eq:agg}
\hspace{-2mm}
\vspace{-2mm}
\end{equation}
where $\bar{a}^m_{e,t}$ is the joint-motion representation of CJT $U^m_t$;
the masked mean makes it invariant to robot ordering. The vector
$q^z_{e,t}$ captures the entity's local interaction context, hypothesized
response mode, and the number of active robots. With
$y_{e,t}=\mathrm{vec}(Y_{e,t})\in\R^{2T}$ and $\mu^{m,z}_{e,t},\sigma^{m,z}_{e,t}\in\R^{2T}$, the FRM uses a mode-conditioned diagonal Gaussian
\vspace{-3mm}

\begin{align}
p_{\phi}\!\left(y_{e,t}\mid \mathcal H_{e,t},U^m_t,z\right) &= \mathcal{N}\!\left(y_{e,t};\ \mu^{m,z}_{e,t},\ \mathrm{diag}\!\left[\left(\sigma^{m,z}_{e,t}\right)^{2}\right]\right),\nonumber\\
\sigma^{m,z}_{e,t} &= \mathrm{softplus}\!\left(\psi^{(\sigma)}\!\left(q^z_{e,t},\bar{a}^m_{e,t}\right)\right),
\label{eq:gauss}
\vspace{-2mm}
\end{align}
where, the diagonal covariance provides a tractable factorized
approximation to the residual uncertainty over the $2T$ velocity
components for likelihood evaluation and risk sampling. The reference JT $U^0_t$ is generated by a fixed rule, encoded by the same network, and its mean $\mu^{0,z}_{e,t}$ is produced by a 
shared baseline decoder $\psi^{(\mu,0)}$.

\noindent\textbf{Effect Decomposition of Conditional Mean.}
The conditional mean used in \eqref{eq:gauss} is decomposed into a reference response, single-robot
effects, and pairwise second-order residuals:
\vspace{-4mm}

\begin{equation}
\hspace{-2mm}
\resizebox{0.45\textwidth}{!}{$
\mu^{m,z}_{e,t} {=} \mu^{0,z}_{e,t}
 {+} \sum_{r=1}^{R^{(\max)}} \hspace{-1mm}\gamma_{r,t}\,\delta^{m,z}_{e,r,t}
{+} \sum_{1\le r<s\le R^{(\max)}}\hspace{-1mm}
\gamma_{r,t}\gamma_{s,t}\,\delta^{m,z}_{e,(r,s),t},$}
\label{eq:anova}
\hspace{-2mm}
\vspace{-1.8mm}
\end{equation}

\noindent where all quantities are vectors in $\mathbb{R}^{2T}$,
corresponding to vectorized $T\times2$ velocity sequences; interactions above second order are omitted. We utilize a multilayer perceptron (MLP) $f_1$ to model the effect of each individual robot, and a second MLP $g$ to capture pairwise robot interactions. To isolate the contribution of robot $r$ relative to its reference motion, we define the centered single-robot effect as follows
\vspace{-1.8mm}

\begin{equation*}
\delta^{m,z}_{e,r,t}
=
f_1\!\left(q^z_{e,t},a^m_{e,r,t}\right)
-
f_1\!\left(q^z_{e,t},a^0_{e,r,t}\right).
\vspace{-2mm}
\end{equation*}
Subsequently, $\delta^{m,z}_{e,r,t}=0$ whenever robot $r$ follows its reference trajectory. To capture the additional response that arises from the joint motion of robots $r$ and $s$, we define
\vspace{-4mm}

\begin{equation*}
G^z_{e,t}(a,b)
=
g\!\left(q^z_{e,t},\rho(a,b)\right),
\
\rho(a,b)
=
[a+b;\,|a-b|;\,a\odot b],
\vspace{-1.5mm}
\end{equation*}
where $\odot$ denotes elementwise multiplication and the construction of $\rho$ is symmetric in its two arguments. The pairwise residual is then defined as
\vspace{-3mm}

\begin{align}
\delta^{m,z}_{e,(r,s),t}
=&\;
G^z_{e,t}\!\left(a^m_{e,r,t},a^m_{e,s,t}\right)
-
G^z_{e,t}\!\left(a^m_{e,r,t},a^0_{e,s,t}\right)
\nonumber\\
&-
G^z_{e,t}\!\left(a^0_{e,r,t},a^m_{e,s,t}\right)
+
G^z_{e,t}\!\left(a^0_{e,r,t},a^0_{e,s,t}\right).
\label{eq:pairres}
\vspace{-2mm}
\end{align}

\noindent This inclusion--exclusion construction removes the effects attributable to either robot alone, so the pairwise residual becomes zero whenever either robot follows its reference trajectory. Consequently, $\delta^{m,z}_{e,(r,s),t}$ isolates the additional mean entity response induced by the joint deviation of robots $r,s$.

\subsubsection{Selection under Predicted Responses}\label{sec:stage1}
The forward coupling uses the FRM predictions to filter and rank the supplied CJTs through the following two steps.

\noindent\textbf{(i) Safety Filtering.}
Given a sampled velocity vector $y_{e,t}$, we first recover the corresponding  T-step velocity sequence
$Y_{e,t}=\mathrm{unvec}(y_{e,t})\in\R^{T\times2}$ and integrate it from the
entity's current position by discrete zero-order-hold accumulation:
\vspace{-2.8mm}

\begin{equation}
p_{e,t,\ell}(Y) = p_{e,t,0} + \Delta t \sum_{\ell'=1}^{\ell} Y_{e,t,\ell'},
\label{eq:pos}
\vspace{-1.5mm}
\end{equation}
where $p_{e,t,0}$ is the current position and $\Delta t$ is the prediction-step duration. Let $p^m_{r,t,\ell}$ be robot $r$'s position at step $\ell$ under CJT $U^m_t$ and consider the $\mathsf{R\&E}$ distance 
\vspace{-1.8mm}

\begin{equation}
d^{m,\ell}_{r,e,t}(Y) = \left\| p^m_{r,t,\ell} - p_{e,t,\ell}(Y) \right\|_2.
\label{eq:dist}
\vspace{-1.5mm}
\end{equation}
We consider the occurrence of a safety violation for entity $e$ if any robot comes within the distance $d^{(\mathrm{hard})}$ at any prediction step. Accordingly, the violation probability of CJT $U_t^m$
is 
\vspace{-4.5mm}

\begin{equation}
\hspace{-2mm}
\resizebox{0.45\textwidth}{!}{$
q^{(\mathrm{viol})}_{e,t}(U_t^m)
{=}\sum_{z\in\Zset}b_{e,t}(z)
\Pr\!\left[
\min_{r\in\mathcal R_t,~1\le\ell\le T}d^{m,\ell}_{r,e,t}(\mathrm{unvec}(y))<d^{(\mathrm{hard})}
\right]\hspace{-.75mm},
$}
\label{eq:viol}
\hspace{-2mm}
\vspace{-1.5mm}
\end{equation}
where $\mathrm{unvec}(\cdot)$ reshapes a vector in $\mathbb{R}^{2T}$ into its corresponding $T\times 2$ velocity sequence.
The posterior $b_{e,t}(z)$ thus weights the violation probability associated with each response mode.
We subsequently derive the safe set of CJTs as 
\vspace{-4mm}

\begin{equation}
\mathcal{C}_t^{(\mathrm{safe})}
=
\left\{
U \in \mathcal{C}_t^{(\mathrm{feas})}
:
q^{(\mathrm{viol})}_{e,t}(U) \le \alpha,\;
\forall e\in\mathcal{E}_t^{(\mathrm{eval})}
\right\},
\label{eq:safeset}
\vspace{-2mm}
\end{equation}
where $\alpha$ is the maximum violation probability, enforcing the constraint in~\eqref{eq:objective}. If no entity requires evaluation, $\mathcal{C}_t^{(\mathrm{safe})}=\mathcal{C}_t^{(\mathrm{feas})}$; if the safe set is empty, the braking fallback is used.

\noindent\textbf{(ii) Cost-based Ranking.}
For a safe CJT, we let $p^m_{r,t,T}$ and $p^0_{r,t,T}$ be the candidate and
reference terminal positions of robot r. We measure goal progress by a cost function:
\vspace{-2.8mm}

\begin{equation}
J^{(\mathrm{goal})}\left( U^m_t \right) = \frac{1}{|\mathcal R_t|} \sum_{r\in\mathcal R_t} \frac{\left\| p^m_{r,t,T} - g_r \right\|_2}{\max\{ \left\| p^0_{r,t,T} - g_r \right\|_2,\varepsilon\}},
\label{eq:goal}
\vspace{-1.9mm}
\end{equation}
where $0<\varepsilon \ll 1$ avoids division by zero.
Smaller values of \eqref{eq:goal} indicate greater progress relative to the reference JT. To distinguish among such safe candidates, let $d^{(\mathrm{soft})}>d^{(\mathrm{hard})}$ denote a preferred separation distance. For a sampled entity response $Y$, we define  the residual interaction risk
\vspace{-5.5mm}

\begin{equation}
J^{(\mathrm{risk})}_{e,t}\left( U^m_t, Y \right) = \max_{r\in\mathcal R_t,~1\le\ell\le T} \left[ d^{(\mathrm{soft})} - d^{m,\ell}_{r,e,t}(Y) \right]_+^2,
\label{eq:risk}
\vspace{-1.2mm}
\end{equation}
where $[x]_+=\max(x,0)$. This quantity penalizes the largest shortfall from the preferred separation distance: taking the maximum prevents a potentially critical near encounter from being obscured by averaging over robots or prediction steps.
We further distinguish CJTs according to the additional motion they induce in an entity relative to the reference JT. Let $v^{m,z}_{e,t,\ell}$ denote the step-$\ell$ velocity obtained from the conditional mean under candidate $m$ and response mode $z$, and define its finite-difference acceleration as
{\small $\dot{v}^{m,z}_{e,t,\ell}=\left(v^{m,z}_{e,t,\ell+1}-v^{m,z}_{e,t,\ell}\right)/\Delta t$} (for $T\ge2$). We define the response burden cost as
\vspace{-4mm}

\begin{equation}
\hspace{-2mm}
\resizebox{0.45\textwidth}{!}{$
J^{(\mathrm{brdn})}_{e,t}\left( U^m_t, z \right) {=} \frac{w_v}{T} \sum_{\ell=1}^{T} \left\| v^{m,z}_{e,t,\ell} - v^{0,z}_{e,t,\ell} \right\|_2^2 \hspace{-1mm}{+} \frac{w_a}{T-1} \sum_{\ell=1}^{T-1} \left\| \dot{v}^{m,z}_{e,t,\ell} - \dot{v}^{0,z}_{e,t,\ell} \right\|_2^2,
$}\hspace{-1.5mm}
\label{eq:burden}
\vspace{-2mm}
\end{equation}
where $w_v$ and $w_a$ weight velocity and acceleration changes. In essence,~\eqref{eq:burden} quantifies the additional entity motion induced by a candidate relative to the reference interaction. 
Let $c_{\mathrm{eval}}=\max(|\mathcal E_t^{(\mathrm{eval})}|,1)$, and let
$\mathbf b_t=\{b_{e,t}:e\in\mathcal E_t^{(\mathrm{eval})}\}$ collect the
beliefs of the evaluated entities. We finally define the overall candidate-ranking cost, i.e., the objective function of \eqref{eq:objective},  as
\vspace{-2.9mm}

\begin{equation}
\hspace{-2mm}
\resizebox{0.45\textwidth}{!}{$
\begin{aligned}
&\mathbb{J}_t\!\left(U_t^m;\mathcal H_t,\mathbf b_t\right)
=\frac{1}{c_{\mathrm{eval}}}\Bigl(\lambda^{(\mathrm{risk})}\!\sum_{e\in\mathcal E_t^{(\mathrm{eval})}}\!
E^m_{e,t}\!\left[
J^{(\mathrm{risk})}_{e,t}\!\left(U_t^m,Y\right)
\right]\\[-1mm]
&{+}\lambda^{(\mathrm{brdn})}\!\sum_{e\in\mathcal E_t^{(\mathrm{eval})}}\!
E^m_{e,t}\!\left[
J^{(\mathrm{brdn})}_{e,t}\!\left(U_t^m,z\right)
\right]
\Bigr){+}\lambda^{(\mathrm{goal})}J^{(\mathrm{goal})}\!\left(U_t^m\right).
\end{aligned}
\vspace{-4mm}
$} \hspace{-2mm}
\label{eq:total}
\vspace{-2.7mm}
\end{equation}
We evaluate
{\footnotesize$\mathbb J_t(U_t^m;\mathcal H_t,\mathbf b_t)$} for each
{\footnotesize$U_t^m\in\mathcal C_t^{(\mathrm{safe})}$} and select the candidate with
the minimum cost as $U_t^\star$, which is desired by \eqref{eq:objective}. 
In \eqref{eq:total}, the risk term is averaged over both the latent
response mode and the mode-conditional predicted entity response:
{\footnotesize
$
E^m_{e,t}\!\left[
J^{(\mathrm{risk})}_{e,t}(U_t^m,Y)
\right]
\triangleq
\sum_{z\in\Zset} b_{e,t}(z)
\mathbb{E}_{y\sim
p_{\phi}(\cdot\mid\mathcal H_{e,t},U_t^m,z)}
\!\left[
J^{(\mathrm{risk})}_{e,t}
\!\left(U_t^m,\mathrm{unvec}(y)\right)
\right].
$
} Also, the response burden term depends only on the latent response mode; thus
{\footnotesize
$
E^m_{e,t}\!\left[
J^{(\mathrm{brdn})}_{e,t}(U_t^m,z)
\right]
\triangleq
\sum_{z\in\Zset} b_{e,t}(z)
J^{(\mathrm{brdn})}_{e,t}(U_t^m,z).
$
}

\subsection{Backward Coupling: Obtaining the Next-Window Belief}\label{sec:backward}
The backward coupling extracts, from the executed part of the selected plan,
the evidence needed to refine the entity-specific belief, and propagates that belief to the next window. It involves two steps: collecting synchronized evidence and updating the identity-bound posterior, detailed below.

\subsubsection{Evidence Collection from the Executed Prefix}\label{sec:stage2}

The system executes the first $L$ steps of $U_t^\star$ through a low-level controller. Because of external disturbances, the realized robot motion may differ from the planned prefix; we therefore record the executed robot states as $U^{(\mathrm{exec})}_{t,1:L}$. Over the same horizon, the system records the synchronized entity velocity response $Y^{(\mathrm{obs})}_{e,t,1:L}$. Together with the entity-local context $\mathcal H_{e,t}$, these observations
form the execution evidence $\xi_{e,t}$ in \eqref{eq:evidence}.

\subsubsection{Posterior Update and Replanning}\label{sec:stage3}
The observed entity response is then used to assess how well each latent response mode $z$ explains the executed interaction. Specifically, for each mode, the FRM evaluates the log-likelihood of the observed response over the executed prefix and obtains
\vspace{-1.8mm}

\begin{equation}
\eta^{(L)}_{e,t}(z) = \log p^{(L)}_{\phi}\left( Y^{(\mathrm{obs})}_{e,t,1:L} \mid \mathcal H_{e,t}, U^{(\mathrm{exec})}_{t,1:L}, z \right),
\label{eq:loglike}
\vspace{-3mm}
\end{equation}
where $p^{(L)}_{\phi}$ is the FRM density restricted to the executed prefix (i.e., the first $L$ prediction steps). The same FRM parameters are used for both prefix and full-horizon prediction. The identity-bound posterior is then updated using Bayes' rule:
\vspace{-4.8mm}

\begin{equation}
\hspace{-4mm}
\resizebox{0.45\textwidth}{!}{$
b_{e,t+1}(z) {=} { b_{e,t}(z) \exp\left( \eta^{(L)}_{e,t}(z) \right) }\Big/{ \sum_{k \in \Zset} b_{e,t}(k) \exp\left( \eta^{(L)}_{e,t}(k) \right) }, \ z \in \Zset.
$}\hspace{-4mm}
\label{eq:bayes}
\vspace{-1.8mm}
\end{equation}
Thus, response modes that better explain the observed interaction receive greater posterior weight. A newly observed entity is initialized with $b_{e,0}(z)$; if no synchronized observation is available in a window, its posterior is carried forward unchanged and is discarded once the identity track terminates. The updated posterior $b_{e,t+1}$ is then used in the next window for response prediction and CJT evaluation, thereby closing the backward-to-forward coupling loop.

\subsection{Offline Training}\label{sec:train}
The FRM must support two related tasks at inference time: predicting full-horizon entity responses for CJT evaluation and assigning likelihoods to shorter executed prefixes for posterior updates. We therefore train both capabilities jointly using the same model parameters. Training samples are generated from paired reference, single-robot-change, and joint-change branches under matched context, scene initial state, exogenous randomness, and response-mode label. 
To train the prefix-likelihood mechanism used in \eqref{eq:bayes}, we sample a prefix length $\tilde{L}\sim P^{(\mathrm{pref})}$, with $\tilde{L}\in\{1,\ldots,T\}$. For training sample $n$, we define the normalized full-horizon and prefix log-likelihoods as 
{\footnotesize $ \ell^{(T)}_n = \frac{1}{2T} \log p^{(T)}_{\phi} \left( Y_{n,1:T} \mid H_n,U_{n,1:T},z_n \right)$} 
and 
{\footnotesize $\ell^{(\mathrm{pref})}_n = \mathbb{E}_{\tilde{L}}\! \Bigl[ \frac{1}{2\tilde{L}} \log p^{(\tilde{L})}_{\phi} ( Y_{n,1:\tilde{L}} \mid H_n,U_{n,1:\tilde{L}},z_n ) \Bigr],$} where 
  {\footnotesize$1/(2T)$} and {\footnotesize$1/(2\tilde{L})$} normalize the log-likelihoods by the two velocity coordinates and the number of time steps. We then define the response-model training objective over a sample batch $\mathcal{B}$ as
\vspace{-1.8mm}

\begin{equation}
\begin{aligned}
\mathcal{L}^{(\mathrm{resp})} ={}& -\frac{1}{|\mathcal{B}|}\sum_{n\in\mathcal{B}}
\Big[\ell^{(T)}_n + \lambda^{(\mathrm{pref})}\ell^{(\mathrm{pref})}_n\Big],
\end{aligned}
\label{eq:loss}
\vspace{-3.0mm}
\end{equation}
where $\lambda^{(\mathrm{pref})}\ge0$ controls the contribution of prefix training. 
We minimize $\mathcal{L}^{(\mathrm{resp})}$ end-to-end with respect to all trainable parameters $\phi$ of the FRM via the gradient descent method.

\section{Experiments}\label{sec:experiments}
We organize the evaluation around the two components of $\mathsf{TRACER}$'s behavior-coupled loop. Specifically, we first isolate the \textit{forward coupling} and ask whether conditioning on joint robot trajectories is necessary to capture non-additive entity responses (Section \ref{Subsec:Sim:B}). We then isolate the \textit{backward coupling} and examine whether evidence gathered from executed interactions can  improve response prediction (Section \ref{Subsec:Sim:C}). Afterwards, we connect these two mechanisms to decision making by testing whether the beliefs improve closed-loop replanning and response-aware navigation (Section \ref{Subsec:Sim:D}). Finally, we evaluate generalization beyond the controlled synthetic setting using recorded human-robot interactions and external navigation benchmarks (Section \ref{Subsec:Sim:E}).


\subsection{Evaluation Protocol}
We use two complementary synthetic protocols to isolate the two components of $\mathsf{TRACER}$'s coupling mechanism. \textit{(i)} The \textit{causal-response protocol} is designed to evaluate the forward model: matched scenes are generated under four robot-motion interventions in which neither, either, or both robot trajectories are changed, allowing us to directly measure non-additive joint effects while holding the initial state and exogenous randomness fixed. \textit{(ii)} The \textit{persistent-mode protocol} is designed to evaluate the backward update: entity response modes remain fixed over multiple replanning windows, allowing us to test whether evidence accumulated from executed interactions improves later predictions and decisions.
In the causal-response protocol, each scene contains 2 robots and 12 entities, with 4,096/256/256 train/validation/test scenes for each of the none, additive, and nonlinear response regimes. In the persistent-mode protocol, each episode contains 3 robots and 10 entities, of which 3 are observed in each window; we use 8,192 training episodes and 256 in-distribution (ID) and 256 out-of-distribution
 (OOD) test episodes, with correlated team actions and fixed response modes over 8 windows. Unless otherwise stated, synthetic experiments use 5 random seeds, 8 history steps, a 12-step prediction horizon,  $\Delta=0.4$ s, a 3-step execution prefix, and model width 96.
We report negative log-likelihood (NLL), average displacement error (ADE), final displacement error (FDE), and paired-effect RMSE where applicable. For synthetic ablations, we report means with standard deviations (SDs) computed across random seeds.

\subsection{Forward Coupling: Joint Actions Recover Non-additive Responses}\label{Subsec:Sim:B}
We first isolate the forward coupling of $\mathsf{TRACER}$ and ask whether an entity's response to multiple robots can be explained by independent single-robot effects, or whether explicitly modeling joint robot motion is necessary. The causal-response protocol is particularly suited to this question because the four matched interventions vary only the robots' candidate motions while keeping the scene initialization and stochastic realization fixed. We compare $\mathsf{TRACER}$'s FRM with five ablations: $\mathsf{HistoryOnly}$ (no robot-action input), $\mathsf{NearestRobot}$ (nearest-robot conditioning), $\mathsf{Additive}$ (no pairwise residual), $\mathsf{MatchedAdditive}$ (capacity-matched additive prediction), and $\mathsf{ShuffledCJT}$ (shuffled candidate inputs). We additionally include protocol-matched $\mathsf{Eiffert\text{-}style}$~\cite{r22}, $\mathsf{ScePT\text{-}style}$~\cite{r13}, and $\mathsf{LatentGame}$ reference adapters. $\mathsf{MatchedAdditive}$ differs from $\mathsf{TRACER}$ by only 0.31\% in residual parameter count, providing a controlled comparison between additive and joint-effect modeling.

\begin{figure}[t]
  \centering
  \includegraphics[width=0.9\columnwidth]{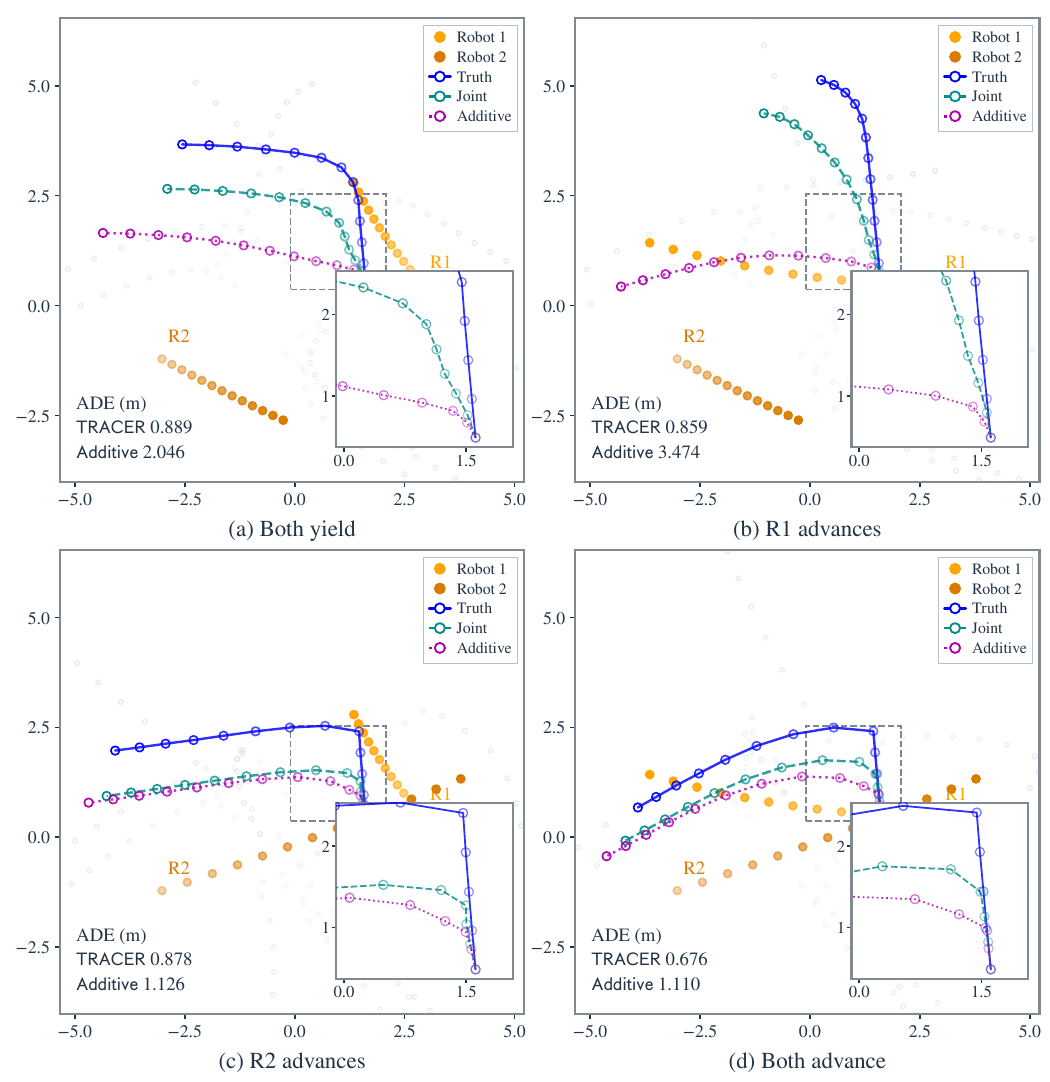}
  \vspace{-4mm}
\caption{Illustrative matched four-intervention scene for exposing a non-additive response. The panels correspond to $Y^{(00)}$ (both robots yield), $Y^{(10)}$ (only R1 advances), $Y^{(01)}$ (only R2 advances), and $Y^{(11)}$ (both robots advance), with the same initial history and exogenous noise. Comparing the single-robot interventions in (b)--(c) with the joint intervention in (d) reveals the response component that cannot be recovered by adding independent robot effects; aggregate performance is reported in Table~\ref{tab:forward_ablation}.}
  \label{fig:joint_trajectories}
  \vspace{-7mm}
\end{figure}

\begin{table}[b]
  \vspace{-5mm}
  \caption{Nonlinear response comparison.
  }
  \label{tab:forward_ablation}
    \vspace{-2mm}
  \centering
  \footnotesize
  \setlength{\tabcolsep}{3pt}
  \input{figures/TRACER_table_forward.tex}
\end{table}

\begin{figure}[!t]
  \centering
  \includegraphics[width=0.9\columnwidth, trim= 0 0 0 15, clip]{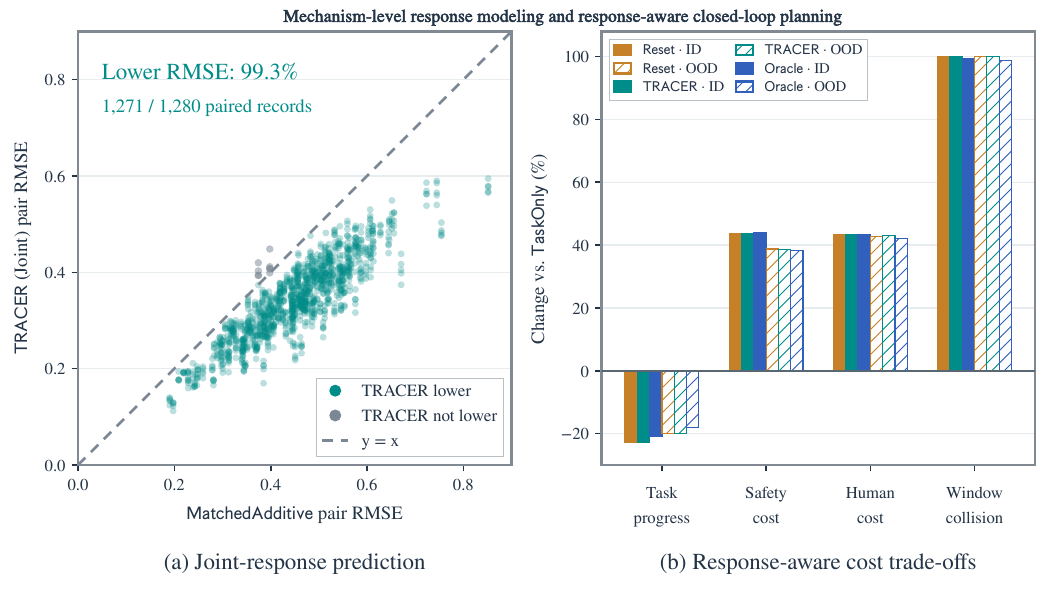}
  \vspace{-5mm}
\caption{Joint-response prediction and response-aware closed-loop planning. (a) Pairwise RMSE comparison between $\mathsf{TRACER}$ and $\mathsf{MatchedAdditive}$ across test scenes and seeds. Each point denotes one record, and points below the $y=x$ line indicate lower error for $\mathsf{TRACER}$. (b) Changes in task progress, safety cost, human-response cost, and collisions compared to $\mathsf{TaskOnly}$ on ID/OOD.}
  \label{fig:response_cost}
  \vspace{-2.5mm}
\end{figure}

To quantify non-additive interaction effects, we compute the second difference $\Delta^{(2)}Y=Y^{(11)}-Y^{(10)}-Y^{(01)}+Y^{(00)}$ which cancels the reference and individual-robot contributions under the matched interventions and isolates the additional response caused by the joint robot-motion change ($Y^{(00)}$, $Y^{(10)}$, $Y^{(01)}$, and $Y^{(11)}$ denote the entity responses when neither robot, only robot 1, only robot 2, and both robots deviate from their reference trajectories, respectively). Pair RMSE measures error in predicting this quantity.
Table~\ref{tab:forward_ablation} shows that explicit joint modeling is particularly important in the nonlinear-response regime. $\mathsf{TRACER}$ achieves the lowest pair RMSE (0.343) and FDE (0.480), reducing pair RMSE by 23.94\% relative to $\mathsf{MatchedAdditive}$. 
$\mathsf{TRACER}$ obtains lower pair RMSE in 1,271 of 1,280 matched
scene-seed records (Fig.~\ref{fig:response_cost}(a)). The qualitative example in Fig.~\ref{fig:joint_trajectories} conveys the same message: when the two robots change their motions, the additive predictor cannot reproduce the resulting entity response, whereas $\mathsf{TRACER}$ captures the interaction effect.


\subsection{Backward Coupling: Retaining Executed Evidence}\label{Subsec:Sim:C}
We next isolate the backward coupling of $\mathsf{TRACER}$. The key question is whether observations from previously executed interactions remain informative in later replanning windows. We therefore fix the FRM and vary only how the response-mode belief is maintained. We compare our method with $\mathsf{NoBelief}$, which uses uniform mode weights, $\mathsf{Reset}$, which discards cross-window evidence, and $\mathsf{ShuffledID}$, which breaks identity associations. As shown in Table~\ref{tab:identity_ablation}, $\mathsf{TRACER}$ achieves the lowest NLL, ADE, and FDE on both ID and OOD splits, showing that retaining evidence across windows provides information beyond the current observation alone. 

\begin{table}[!b]
\vspace{-4mm}
  \caption{Evidence-retention ablations of $\mathsf{TRACER}$ with a fixed FRM: mean $\pm$ seed SD. Bold: best listed mean per split.}
  \label{tab:identity_ablation}
   \vspace{-2mm}
  \centering
  \footnotesize
  \setlength{\tabcolsep}{3pt}
  \input{figures/TRACER_table_identity.tex}
\end{table}

\begin{figure}[!t]
  \centering
  \includegraphics[width=\columnwidth, trim= 0 0 0 16, clip]{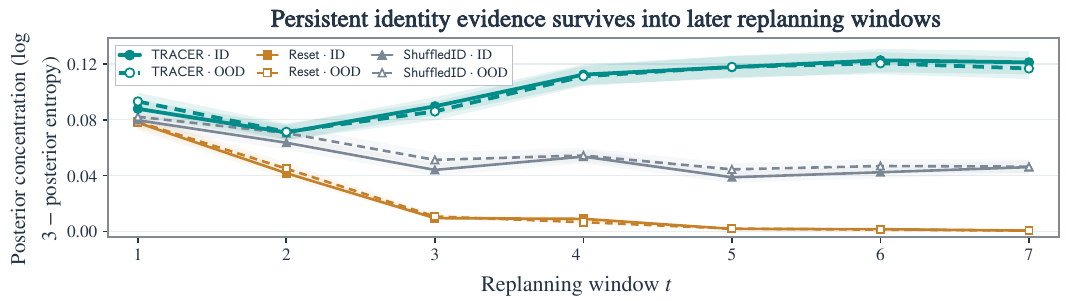}
  \vspace{-8mm}
\caption{Posterior concentration across replanning windows for $\mathsf{TRACER}$, $\mathsf{Reset}$, and $\mathsf{ShuffledID}$ on the ID and OOD splits. Higher values indicate more concentrated identity-bound beliefs.}
  \label{fig:posterior_concentration}
  \vspace{-7mm}
\end{figure}

Fig.~\ref{fig:posterior_concentration} provides a complementary view of this phenomenon. $\mathsf{TRACER}$'s posterior remains progressively concentrated across later replanning windows, whereas $\mathsf{Reset}$ returns toward a nearly uniform belief after evidence is discarded. Together, these results show that executed interaction evidence remains useful across planning windows, but only when it is persistently associated with the correct entity.

\subsection{From Belief to Action: Evidence-Informed Replanning}\label{Subsec:Sim:D}
The previous experiment shows that persistent interaction evidence improves
response prediction. We next examine how this improvement translates into
closed-loop navigation decisions. As shown in
Fig.~\ref{fig:response_cost}(b), we compare $\mathsf{TRACER}$ with
$\mathsf{TaskOnly}$, which optimizes task progress alone,
$\mathsf{Reset}$, which uses the same response model but discards
cross-window evidence, and $\mathsf{Oracle}$, which has privileged access
to the true response mode. Relative to $\mathsf{TaskOnly}$, $\mathsf{TRACER}$ substantially reduces
both safety and human-response costs on the ID and OOD splits, while
incurring a moderate loss in task progress. Moreover, the gap between
$\mathsf{TRACER}$ and $\mathsf{Reset}$ shows that retaining
identity-consistent evidence across replanning windows improves the
resulting response-aware decisions. The performance of $\mathsf{Oracle}$
provides an upper reference corresponding to perfect knowledge of the
latent response mode. Overall, these results show that the backward belief
update of $\mathsf{TRACER}$ affects downstream candidate selection and helps achieve a better
balance between task progress and interaction-aware navigation.

\subsection{Generalization Beyond Controlled Simulation}\label{Subsec:Sim:E}
\begin{figure}[!tb]
  \centering
  \includegraphics[width=0.9\columnwidth]{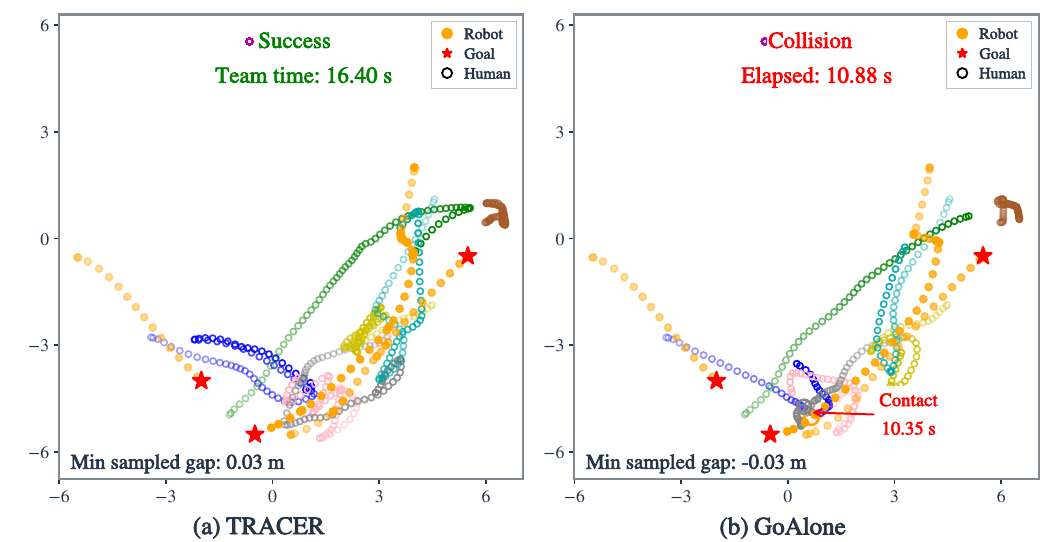}
  \vspace{-4mm}
\caption{Executions from matched initial states under $\mathsf{TRACER}$ and $\mathsf{GoAlone}$. (a) $\mathsf{TRACER}$ completes the task while maintaining a positive minimum sampled gap. (b) $\mathsf{GoAlone}$ finally enters contact under the same initial state.}
  \label{fig:external_trajectories}
  \vspace{-7mm}
\end{figure}

The preceding experiments study $\mathsf{TRACER}$ under controlled conditions. We finally examine whether the learned response model and the resulting closed-loop navigation behavior remain effective outside those synthetic protocols. For clarity, $\mathsf{SICNav\text{-}P}$ and $\mathsf{SICNav\text{-}NP}$ are the privileged and non-privileged $\mathsf{SICNav}$ variants, while $\mathsf{MPC\text{-}CVMM}$ replaces $\mathsf{SICNav}$'s $\mathsf{ORCA}$ response model with a constant-velocity model~\cite{r3}. $\mathsf{DWA}$~\cite{r44} is a reactive unicycle baseline; $\mathsf{ORCA}$~\cite{r17} and $\mathsf{Linear}$ are holonomic references. In SocialGym2, $\mathsf{GoAlone}$ and $\mathsf{Halt}$ are official advance/stop controls~\cite{socialgym2}. The $\mathsf{TRACER}$ variants remove pairwise residuals ($\mathsf{NoPair}$), use uniform beliefs ($\mathsf{NoBelief}$), or discard history ($\mathsf{Reset}$); full $\mathsf{TRACER}$ retains identity-bound evidence.

\begin{table}[!ht]
\vspace{-3mm}
  \caption{CF/Team: collision-free/team completion; Coll.: collision episodes; Prog.: progress; rates in \%. $^{\dagger}$: privileged; $^{\ddagger}$: holonomic.}
  \footnotesize
  \label{tab:closed_loop_ablation}
  \vspace{-4mm}
  \centering
  \input{figures/TRACER_external_v19.tex}
  \vspace{-4mm}
\end{table}

\textbf{(i) Recorded Human--Robot Interactions.} We first evaluate whether action conditioning remains beneficial on
recorded human-robot interaction data: NavWareSet~\cite{navwareset} and the complete Bi$^3$ dataset~\cite{bi3} contains 26/8/9 and 135/25/25 train/validation/test recordings, respectively, with up to two humans. Nine observations precede a 12-step prediction horizon (1.2/4.8~s); both predictors use five-epoch refinement with three seeds. As reported in Table~\ref{tab:recorded_prediction}, action conditioning improves NLL, ADE, and FDE over $\mathsf{HistoryOnly}$ on both datasets.

\begin{table}[!t]
  \vspace{-3mm}
  \caption{Recorded interactions: test-split means over three seeds.}
  \label{tab:recorded_prediction}
  \vspace{-3mm}
  \centering
  \footnotesize
  \setlength{\tabcolsep}{3pt}
  \input{figures/TRACER_table_recorded.tex}
  \vspace{-7mm}
\end{table}

\textbf{(ii) External Closed-Loop Navigation.} We finally evaluate the complete navigation framework in CrowdSimPlus and SocialGym2. These experiments test whether the response-modeling and adaptation mechanisms translate into collision-free multi-robot navigation under external simulator dynamics.
Table~\ref{tab:closed_loop_ablation} summarizes the results: on CrowdSimPlus, $\mathsf{TRACER}$ achieves 100\% collision-free completion, matching the privileged $\mathsf{SICNav\text{-}P}$ baseline and outperforming $\mathsf{SICNav\text{-}NP}$ and $\mathsf{MPC\text{-}CVMM}$. It also achieves a progress score of 0.951, the highest among the kinematically matched non-privileged controllers. On SocialGym2, which contains 320 paired evaluation episodes with three robots and two humans, $\mathsf{TRACER}$ improves collision-free completion from 22.5\% to 31.6\% and team completion from 40.9\% to 47.5\% relative to the independent-robot $\mathsf{GoAlone}$ baseline, while reducing the collision-episode rate from 53.8\% to 30.3\%. 
Fig.~\ref{fig:external_trajectories} complements these aggregate results with an example in which $\mathsf{TRACER}$ avoids the contact incurred by $\mathsf{GoAlone}$.


\vspace{-2mm}
\section{Conclusion}\label{sec:conclusion}
In this work, we proposed $\mathsf{TRACER}$, a closed-loop framework for multi-robot social navigation in which robot plans shape entity responses and executed interactions inform future decisions. By combining joint-trajectory-conditioned response prediction with persistent, identity-bound belief updates, $\mathsf{TRACER}$ captures non-additive interaction effects and adapts subsequent replanning from observed outcomes. Experiments showed that joint multi-robot interaction modeling improves response prediction, while persistent identity-consistent evidence refines future predictions and downstream navigation decisions. Evaluations on external benchmarks further demonstrated improved collision-aware multi-robot navigation relative to competing baselines. Together, these results show that closing the loop between predicted responses, executed interactions, and future planning provides an effective basis for socially-aware multi-robot navigation.

\vspace{-2mm}


\scriptsize{
\bibliographystyle{IEEEtran}
\bibliography{references}}

\end{document}

%% file: figures/TRACER_table_forward.tex
\begin{tabular}{@{}lrrr@{}}
\noalign{\hrule height 0.7pt}
\noalign{\begingroup\color[gray]{0.88}\hrule height\dimexpr\ht\strutbox+\dp\strutbox\relax\endgroup\vskip-\dimexpr\ht\strutbox+\dp\strutbox\relax}
\textbf{Predictor} & \textbf{ADE $\downarrow$} & \textbf{FDE $\downarrow$} & \textbf{Pair RMSE $\downarrow$} \\
\hline
$\mathsf{HistoryOnly}$ & $1.926\,\pm\,0.002$ & $2.807\,\pm\,0.004$ & $0.451\,\pm\,0.000$ \\
$\mathsf{NearestRobot}$ & $0.760\,\pm\,0.003$ & $1.271\,\pm\,0.005$ & $0.451\,\pm\,0.000$ \\
$\mathsf{Additive}$ & $0.571\,\pm\,0.006$ & $0.741\,\pm\,0.008$ & $0.451\,\pm\,0.000$ \\
$\mathsf{MatchedAdditive}$ & $0.518\,\pm\,0.012$ & $0.681\,\pm\,0.011$ & $0.451\,\pm\,0.000$ \\
$\mathsf{ShuffledCJT}$ & $1.363\,\pm\,0.007$ & $2.361\,\pm\,0.007$ & $0.952\,\pm\,0.007$ \\
\hline
$\mathsf{Eiffert\text{-}style}$ & $0.472\,\pm\,0.013$ & $0.589\,\pm\,0.013$ & $0.398\,\pm\,0.002$ \\
$\mathsf{ScePT\text{-}style}$ & $\mathbf{0.379}\,\pm\,0.002$ & $0.566\,\pm\,0.002$ & $0.378\,\pm\,0.001$ \\
$\mathsf{LatentGame}$ & $0.446\,\pm\,0.012$ & $0.561\,\pm\,0.013$ & $0.397\,\pm\,0.002$ \\
\hline
$\bm{\mathsf{TRACER}}$ & $0.402\,\pm\,0.015$ & $\mathbf{0.480}\,\pm\,0.017$ & $\mathbf{0.343}\,\pm\,0.003$ \\
\noalign{\hrule height 0.7pt}
\end{tabular}

%% file: figures/TRACER_table_identity.tex
\begingroup
\renewcommand{\arraystretch}{1.10}
\setlength{\tabcolsep}{1.75pt}
\begin{tabular*}{\columnwidth}{@{\extracolsep{\fill}}llrrr@{}}
\noalign{\hrule height 0.7pt}
\noalign{\begingroup\color[gray]{0.90}\hrule height\arraystretch\dimexpr\ht\strutbox+\dp\strutbox\relax\vskip-\arraystretch\dimexpr\ht\strutbox+\dp\strutbox\relax\endgroup}
\textbf{Split} & \textbf{Variant} & \textbf{NLL $\downarrow$} & \textbf{ADE $\downarrow$} & \textbf{FDE $\downarrow$} \\
\noalign{\hrule height 0.4pt\vskip 1.5pt}
\textbf{ID} & $\mathsf{NoBelief}$ & $-0.0340\,\pm\,0.0030$ & $0.644\,\pm\,0.009$ & $1.117\,\pm\,0.011$ \\
 & $\mathsf{Reset}$ & $-0.0362\,\pm\,0.0031$ & $0.623\,\pm\,0.006$ & $1.084\,\pm\,0.008$ \\
 & $\mathsf{ShuffledID}$ & $-0.0324\,\pm\,0.0029$ & $0.648\,\pm\,0.008$ & $1.124\,\pm\,0.010$ \\
 & $\bm{\mathsf{TRACER}}$ & $\mathbf{-0.0402}\,\pm\,0.0031$ & $\mathbf{0.599}\,\pm\,0.004$ & $\mathbf{1.037}\,\pm\,0.007$ \\
\noalign{\vskip 2pt\hrule height 0.35pt\vskip 2pt}
\textbf{OOD} & $\mathsf{NoBelief}$ & $-0.0404\,\pm\,0.0036$ & $0.641\,\pm\,0.008$ & $1.109\,\pm\,0.010$ \\
 & $\mathsf{Reset}$ & $-0.0426\,\pm\,0.0036$ & $0.620\,\pm\,0.006$ & $1.074\,\pm\,0.009$ \\
 & $\mathsf{ShuffledID}$ & $-0.0385\,\pm\,0.0035$ & $0.646\,\pm\,0.007$ & $1.117\,\pm\,0.010$ \\
 & $\bm{\mathsf{TRACER}}$ & $\mathbf{-0.0466}\,\pm\,0.0037$ & $\mathbf{0.595}\,\pm\,0.005$ & $\mathbf{1.025}\,\pm\,0.009$ \\
\noalign{\vskip 1.5pt\hrule height 0.7pt}
\end{tabular*}
\endgroup

%% file: figures/TRACER_external_v19.tex
\textbf{(a) CrowdSimPlus: 24 paired development scenes}\par\smallskip
\begingroup
\renewcommand{\arraystretch}{0.92}
\setlength{\tabcolsep}{2pt}
\begin{tabular*}{\columnwidth}{@{\extracolsep{\fill}}lrrr@{}}
\noalign{\hrule height 0.7pt}
\noalign{\begingroup\color[gray]{0.90}\hrule height\arraystretch\dimexpr\ht\strutbox+\dp\strutbox\relax\vskip-\arraystretch\dimexpr\ht\strutbox+\dp\strutbox\relax\endgroup}
\textbf{Controller} & \textbf{CF $\uparrow$} & \textbf{Time $\downarrow$} & \textbf{Prog. $\uparrow$} \\
\noalign{\hrule height 0.4pt\vskip 1pt}
$\mathsf{SICNav\text{-}NP}$ & $95.8$ & $4.05$ & $0.938$ \\
$\mathsf{SICNav\text{-}P}^{\dagger}$ & $100.0$ & $4.04$ & $0.934$ \\
$\mathsf{MPC\text{-}CVMM}$ & $91.7$ & $4.49$ & $0.941$ \\
$\mathsf{DWA}$ & $45.8$ & $11.76$ & $0.619$ \\
$\mathsf{ORCA}^{\ddagger}$ & $58.3$ & $8.41$ & $0.774$ \\
$\mathsf{Linear}^{\ddagger}$ & $66.7$ & $3.00$ & $1.000$ \\
$\bm{\mathsf{TRACER}}$ & $100.0$ & $4.39$ & $0.951$ \\
\noalign{\vskip 1pt\hrule height 0.7pt}
\end{tabular*}
\endgroup
\par\vspace{1pt}\textbf{(b) SocialGym2: 320 paired evaluation episodes}\par\vspace{1pt}
\begingroup
\renewcommand{\arraystretch}{0.92}
\setlength{\tabcolsep}{2pt}
\begin{tabular*}{\columnwidth}{@{\extracolsep{\fill}}lrrrr@{}}
\noalign{\hrule height 0.7pt}
\noalign{\begingroup\color[gray]{0.90}\hrule height\arraystretch\dimexpr\ht\strutbox+\dp\strutbox\relax\vskip-\arraystretch\dimexpr\ht\strutbox+\dp\strutbox\relax\endgroup}
\textbf{Policy} & \textbf{CF $\uparrow$} & \textbf{Team $\uparrow$} & \textbf{Coll. $\downarrow$} & \textbf{Prog. $\uparrow$} \\
\noalign{\hrule height 0.4pt\vskip 1pt}
$\mathsf{GoAlone}$ & $22.5$ & $40.9$ & $53.8$ & $0.805$ \\
$\mathsf{Halt}$ & $0.0$ & $0.0$ & $26.9$ & $0.000$ \\
$\mathsf{TRACER}(\mathsf{NoPair})$ & $31.9$ & $48.1$ & $30.0$ & $0.840$ \\
$\mathsf{TRACER}(\mathsf{NoBelief})$ & $32.2$ & $49.1$ & $29.4$ & $0.839$ \\
$\mathsf{TRACER}(\mathsf{Reset})$ & $32.2$ & $48.8$ & $29.7$ & $0.837$ \\
$\bm{\mathsf{TRACER}}$ & $31.6$ & $47.5$ & $30.3$ & $0.839$ \\
\noalign{\vskip 1pt\hrule height 0.7pt}
\end{tabular*}
\endgroup

%% file: figures/TRACER_table_recorded.tex
\begingroup
\renewcommand{\arraystretch}{0.94}
\begin{tabular*}{\columnwidth}{@{\extracolsep{\fill}}llrrr@{}}
\noalign{\hrule height 0.7pt}
\noalign{\begingroup\color[gray]{0.90}\hrule height\arraystretch\dimexpr\ht\strutbox+\dp\strutbox\relax\vskip-\arraystretch\dimexpr\ht\strutbox+\dp\strutbox\relax\endgroup}
\textbf{Dataset} & \textbf{Predictor} & \textbf{NLL $\downarrow$} & \textbf{ADE $\downarrow$} & \textbf{FDE $\downarrow$} \\
\noalign{\hrule height 0.4pt}
NavWareSet & History-only & $-0.8051$ & $0.0592$ & $0.1267$ \\
NavWareSet & Action-conditioned & $\mathbf{-0.8127}$ & $\mathbf{0.0581}$ & $\mathbf{0.1240}$ \\
Bi$^3$ & History-only & $0.1700$ & $0.4457$ & $0.6686$ \\
Bi$^3$ & Action-conditioned & $\mathbf{0.1566}$ & $\mathbf{0.4369}$ & $\mathbf{0.6598}$ \\
\noalign{\hrule height 0.7pt}
\end{tabular*}
\endgroup